\documentclass[journal]{IEEEtran}
\ifCLASSINFOpdf
\else
\fi
\usepackage{ifpdf}
\ifpdf
  \usepackage{epstopdf}
\fi

\usepackage{times,amsmath,epsfig}
\usepackage{epstopdf}
\usepackage{graphicx}
\usepackage{cite}
\usepackage{array}
\usepackage{amsmath}
\usepackage{booktabs}
\usepackage{multirow}
\usepackage{multicol}
\usepackage{algorithm}
\usepackage{algorithmicx}
\usepackage{algpseudocode}
\usepackage{url}
\usepackage{diagbox}
\usepackage{color}
\usepackage[normalem]{ulem}

\newcommand{\norm}[1]{\left\lVert#1\right\rVert}

\begin{document}
%
\title{An End-to-End Compression Framework Based on Convolutional Neural Networks}
%
%
%
\author{Feng~Jiang,
        Wen~Tao,
        Shaohui~Liu,
        Jie Ren,
        Xun Guo,
        Debin~Zhao,~\IEEEmembership{~Member,~IEEE}
\thanks {F. Jiang, W. Tao, S. Liu and D. Zhao are with the School of Computer Science and Technology, Harbin Institute of Technology, Harbin 150001, China (e-mail: shliu@hit.edu.cn).}}

%
%

\markboth{Submitted to IEEE TRANSACTIONS ON Circuits and Systems for Video Technology}
{Shell \MakeLowercase{\textit{et al.}}: Bare Demo of IEEEtran.cls for IEEE Journals}
%



\maketitle

\begin{abstract}
Deep learning, e.g., convolutional neural networks (CNNs), has achieved great success in image processing and computer vision especially in high level vision applications such as recognition and understanding. However, it is rarely used to solve low-level vision problems such as image compression studied in this paper. Here, we move forward a step and propose a novel compression framework based on CNNs. To achieve high-quality image compression at low bit rates, two CNNs are seamlessly integrated into an end-to-end compression framework. The first CNN, named compact convolutional neural network (ComCNN), learns an optimal  compact representation from an input image, which preserves the structural information and is then encoded using an image codec (e.g., JPEG, JPEG2000 or BPG). The second CNN, named reconstruction convolutional neural network (RecCNN), is used to reconstruct the decoded image with high-quality in the decoding end. To make two CNNs effectively collaborate, we develop a unified end-to-end learning algorithm to simultaneously learn ComCNN and RecCNN, which facilitates the accurate reconstruction of the decoded image using RecCNN. Such a design also makes the proposed compression framework compatible with existing image coding standards. Experimental results validate that the proposed compression framework greatly outperforms several compression frameworks that use existing image coding standards with state-of-the-art deblocking or denoising post-processing methods.
\end{abstract}


%
\IEEEpeerreviewmaketitle

\section{Introduction}
%
%
%
%
\IEEEPARstart{I}{n} recent years, image compression attracts increasing interest in image processing and computer vision due to its potential applications in many vision systems. The aim of image compression is to reduce irrelevance and redundancy of an image in order to store or transmit the image at low bit rates \cite{wallace1992jpeg}. Traditional image coding standards\cite{ghanbari2003standard} (such as JPEG and JPEG2000) attempt to distribute the available bits for every nonzero quantized transform coefficient in the whole image. While the compression ratio increases, the bits per pixel (BPP) decreases as a result of the use of bigger quantization steps, which will cause the decoded image to have blocking artifacts or noises. To overcome this problem, a lot of efforts have been devoted to improving the quality of the decoded image using a post-processing deblocking or denoising method. Zhai et al. \cite{zhai2008efficient} propose an effective deblocking method for JPEG images through post-filtering in shifted windows of image blocks. Foi et al. \cite{foi2007pointwise} develop an image deblocking filtering based on shape-adaptive DCT, in conjunction with the anisotropic local polynomial approximation-intersection of confidence intervals technique. Inspired by the success of nonlocal filters and bilateral filters for image debolcking, several nonlocal filters have been proposed for image deblocking \cite{zhang2011image,francisco2012generic,wang2013adaptive}. Recently, Zhang et al. \cite{zhang2016concolor} propose a constrained non-convex low-rank model for image deblocking. Although desired performance is achieved, these post-processing methods are very time-consuming because solving the optimal solutions involves computationally expensive iterative processes. Therefore, it is difficult to apply them to practical applications.

\begin{figure}
\centering
\includegraphics[width=0.48\textwidth]{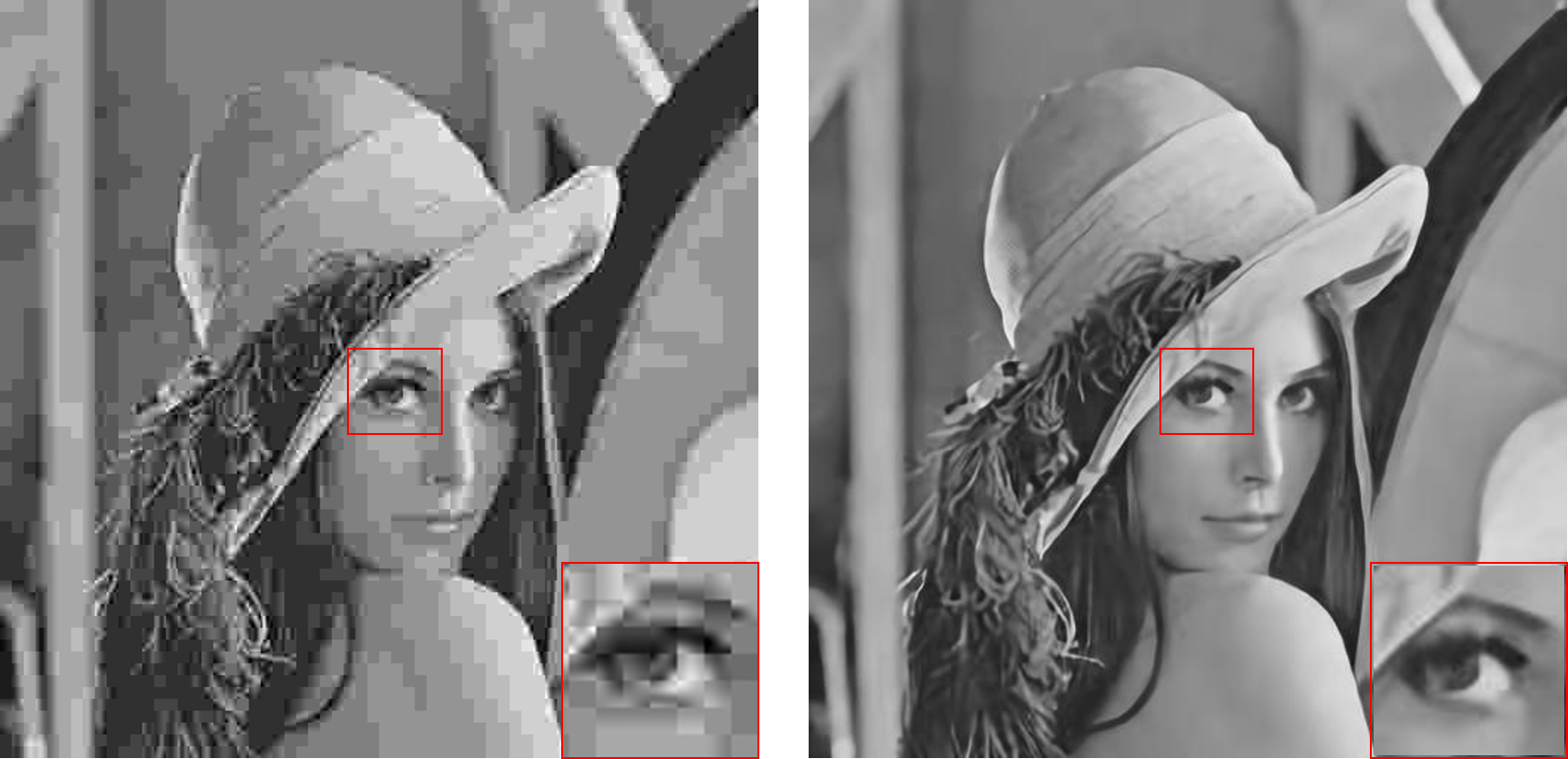}\\
\caption{Left: the JPEG-coded image (PSNR = 27.33 dB) with QF = 5 , where we could see blocking artifacts,ring effects and blurring on the eyes, abrupt intensity changes on the face. Right: the decoded image (PSNR = 31.14 dB) by our proposed compression framework at the same bit rate with left, where the compression artifacts vanished and generate more details.}
\label{Lena}
\end{figure}

\begin{figure*}
\centering
\includegraphics[width=0.7\textwidth]{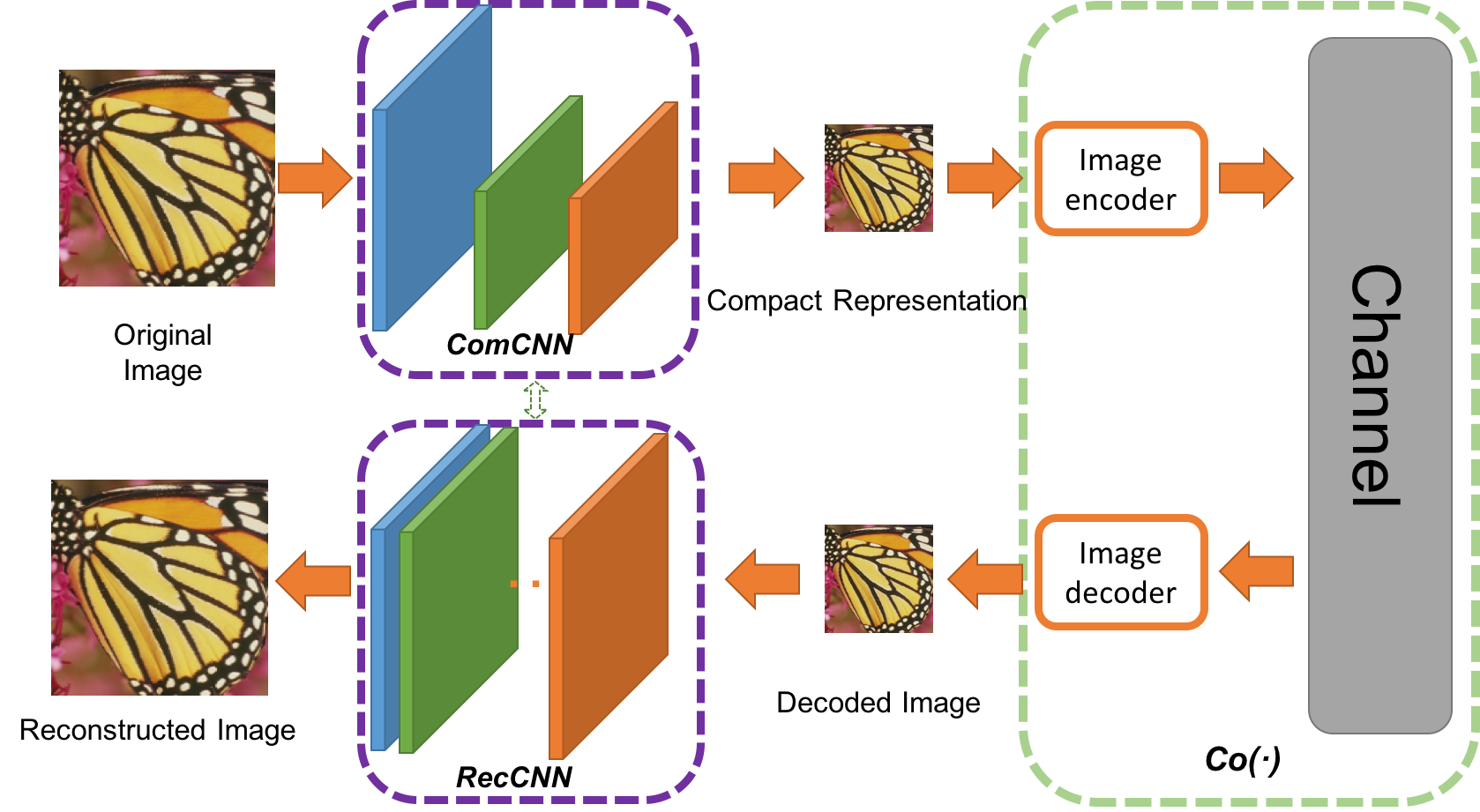}\\
\caption{ The proposed novel compression framework. The ComCNN preserves more useful information for reconstruction. Meanwhile, the RecCNN handle the task of reconstructing the decoded image. $Co(\cdot)$ represents an image codec, encoding and decoding the compact representation  produced by the ComCNN, which consists of a transform coding stage followed by quantization and entropy coding. In this work, JPEG, JPEG2000 and BPG are adopted.}
\label{framework}
\end{figure*}

Motivated by the excellent performance of convolutional neural networks (CNNs) in low level computer vision \cite{dong2015compression, guo2016building, wang2016d3} in recent years and the fact that existing image codecs are extensively used across the world, we propose an end-to-end compression framework,   which consists of two CNNs and an image codec as shown in Fig. 2. The first CNN, named compact convolutional neural network (ComCNN), learns an optimal compact representation from an input image, which is then
encoded using an image codec (e.g., JPEG, JPEG2000 or BPG). The second CNN, named reconstruction convolutional neural network (RecCNN), is used to reconstruct the decoded image with high quality in the decoding end. Existing image coding standards usually consists of transformation, quantization and entropy coding. Unfortunately, the rounding function in quantization is not differentiable, which brings great challenges to train deep neural networks when performing the backpropagation algorithm. To address this problem, we present a simple but effective learning algorithm to train the proposed end-to-end compression framework by simultaneously learning ComCNN and RecCNN to facilitate the accurately reconstruction of the decoded image using RecCNN.  An example of image compression is shown in Fig.~\ref{Lena}, from which we can see that the proposed framework achieves much better quality with more visual details. In addition, as shown in Fig.~\ref{framework}, the compact representation obtained by ComCNN preserves the structural information of the image, therefore, an image codec can be effectively utilzed to compress the compact representation.

The contributions of this work are summarized as follows:
\begin{enumerate}

\item We propose an end-to-end compression framework using two CNNs and an image codec. ComCNN produces a compact representation for encoding using an image codec. RecCNN reconstructs the decoded image, respectively. To our best knowledge, it is the first time to connect the existing image coding standards with CNNs using a compact intermediate representation.

\item We further propose an effective learning algorithm to simultaneously learn two CNNs, which addresses the problem that the gradient can not be passed in the backpropagation algorithm since the rounding function in quantization is not differentiable.

\item The proposed compression framework is compatible with existing image codecs (e.g. JPEG, JPEG2000 or BPG), which makes our method more applicable to other tasks.

\end{enumerate}

The remainder of this paper is organized as follows. Section 2 presents a brief review of related work. Section 3 elaborates the proposed compression framework, including the architectures of ComCNN and RecCNN. Section 4 illustrates the training parameters setting and the solutions to train the ComCNN and RecCNN. Experimental results are also reported in Section 4. In Section 5, we conclude this paper.

\section{Related Work}
\subsection{Image Deblocking and Artifacts Reduction}
In the literature, there have been some methods proposed to improve the quality of decoded images using post-processing techniques, which can be roughly categorized into deblocking oriented and restoration oriented methods. The deblocking oriented methods focus on removing blocking and ringing artifacts of the decoded images. Yeh et al.\cite{yeh2014self} propose a self-learning based post-processing method for image/video deblocking by formulating deblocking as a morphological component analysis based image decomposition problem. Yoo et al.\cite{yoo2014post} propose a two-step framework for reducing blocking artifacts in different regions based on increment of inter-block correlation, which classifies the coded image into flat regions and edge regions. Liu et al.\cite{liu2016data} learned sparse representations within the dual DCT-pixel domain, and achieved very promising results. Recently, Dong et al.\cite{dong2015compression} propose a compact and efficient network (ARCNN) for seamless attenuation of different compression artifacts. Innovatively, D3\cite{wang2016d3} and DDCN\cite{guo2016building} integrate dual-domain sparse coding and the prior knowledge of JPEG compression, which achieve impressive results.

The restoration oriented methods regard the compression operation as a distortion process and reduce artifacts by restoring the clear images. Sun et al. \cite{sun2007postprocessing} model the quantization distortion as Gaussian noises and use field of experts as image priors to restore the images. Zhang et al. \cite{zhang2012reducing,zhang2013compression} propose to utilize similarity priors of image blocks to reduce compression artifacts by estimating the transform coefficients of overlapped blocks from non-local blocks. Recently, Zhang et al.\cite{zhang2016concolor} develop a novel algorithm for image deblocking using a constrained non-convex low-rank model, which formulates image deblocking as an optimization problem within maximum a posteriori framework.

In the aforementioned methods, image prior models play important roles in both the deblocking oriented and restoration oriented methods. However, these methods involve computationally expensive iterative processes when solving the optimal solutions with complex formula derivations. Therefore, they may be not suitable for practical applications. In short, all the related methods reviewed above attempt to improve image quality only from the perspective of image post-processing. In other words, the connection between the encoder front-end processing and the decoder back-end processing is ignored. We attempt to jointly optimize the encoder and decoder joint optimization to improve the compression performance.

\subsection{Image Super-Resolution Based on Deep Learning}
Recently, CNNs have been used successfully for image super-resolution (SR) especially when residual learning\cite{he2015deep} and gradients-based optimization algorithms\cite{duchi2011adaptive,zeiler2012adadelta,kingma2014adam} are proposed to train deeper network efficiently. Dong et al. propose a CNN based SR method\cite{dong2016image} named SRCNN, which consists of three layers: patch extraction, non-linear mapping and reconstruction. Although Dong et al. conclude in their paper that deeper networks do not result in better performance in some cases, other researchers argue that increasing depth significantly boosts performance. For example, VDSR\cite{kim2015accurate} shows a significant improvement in accuracy, which uses 20 weight layers. DRCN\cite{kim2015deeply} has a very deep recursive layer (up to 16 recursions) and outperforms previous methods by a large margin.
	
\subsection{Image Compression Based on Deep Learning}
Recently, deep learning has been used both for lossy and lossless image compression and achieved competitive performance. For the lossy image compression,
Toderici et al.\cite{toderici2015variable} propose a general framework for variable-rate image compression and a novel architecture based on convolutional and deconvolutional LSTM  recurrent networks. Further, Toderici et al.\cite{toderici2016full} proposed a neural network which is competitive across compression rates on images of arbitrary size. For a given compression rate, both methods learn the compression models by minimizing the distortion. Theis et al.\cite{theis2017lossy} propose compressive autoencoders, which uses a smooth approximation of the discrete of the rounding function and upper-bound the discrete entropy rate loss for continuous relaxation. Ball\'e et al.\cite{balle2016end} make use a generalized divisive normalization (GDN) for joint nonlinearity and replace rounding quantization with additive uniform noise for continuous relaxation. Li et al.\cite{li2017learning} proposed a content-weighted compression method with the importance map of image. For the lossless image compression, the methods proposed by Theis et al.\cite{theis2015generative} and van den Oord et al.\cite{oord2016pixel} achieves state-of-the-art results. 

Overall, although the image compression methods based on deep learning achieve competitive performance, they ignored the compatibility with existing image codecs, which limits their use in some existing systems. In contrast, the proposed compression framework takes into account both compression performance and compatibility with existing image codecs.

\section{The Proposed Compression Framework}
In this section, we first introduce the architecture of the proposed compression framework and then present the detailed learning algorithm.

\subsection{Architecture of End-to-End Compression Framework}
As shown in Fig.~\ref{framework}, the proposed compression framework consists of two CNNs and an image codec. The compact representation CNN (ComCNN) is used to generate a compact representation of the input image for the encoding, which preserves structural information of the image and therefore facilitates the accurate reconstruction of high-quality images. The reconstruction CNN (RecCNN) is used to enhance the quality of the decoded image. These two CNNs collaborate with each other and are optimized simultaneously to achieve high-quality image compression at low bit rates.

\subsubsection{Compact Representation Convolutional Neural Network (ComCNN)}

As shown in Fig.~\ref{ComCNN}, ComCNN has 3 weight layers, which maintain the spatial structure of the original image and therefore facilitate the accurate reconstruction of the decoded image using RecCNN{\footnote{We have tried deeper networks to obtain better performance, but only negligible improvements at the expense of a lot of training time and costs. }. The combination of convolution and ReLU \cite{krizhevsky2012imagenet} is used in ComCNN. The first layer is used to perform patch extraction and representation which extracts overlapping patches from the input image. Let $c$ represents the number of image channels. A total of 64 filters of size $3\times 3\times c$  are used to generate 64 feature maps and the ReLU nonlinearity is utilized as an activation function. The second layer has two significant intentions: downscaling and enhancing the features, which are implemented by convolutional operations with setting the stride to 2. The sizes of 64 filters are $3\times 3\times 64$ and ReLU is also applied. For the last layer, c filters of size $3\times 3\times 64$ are utilized to construct the compact representation.

\begin{figure*}
\centering
\includegraphics[width=0.7\textwidth, height=5cm]{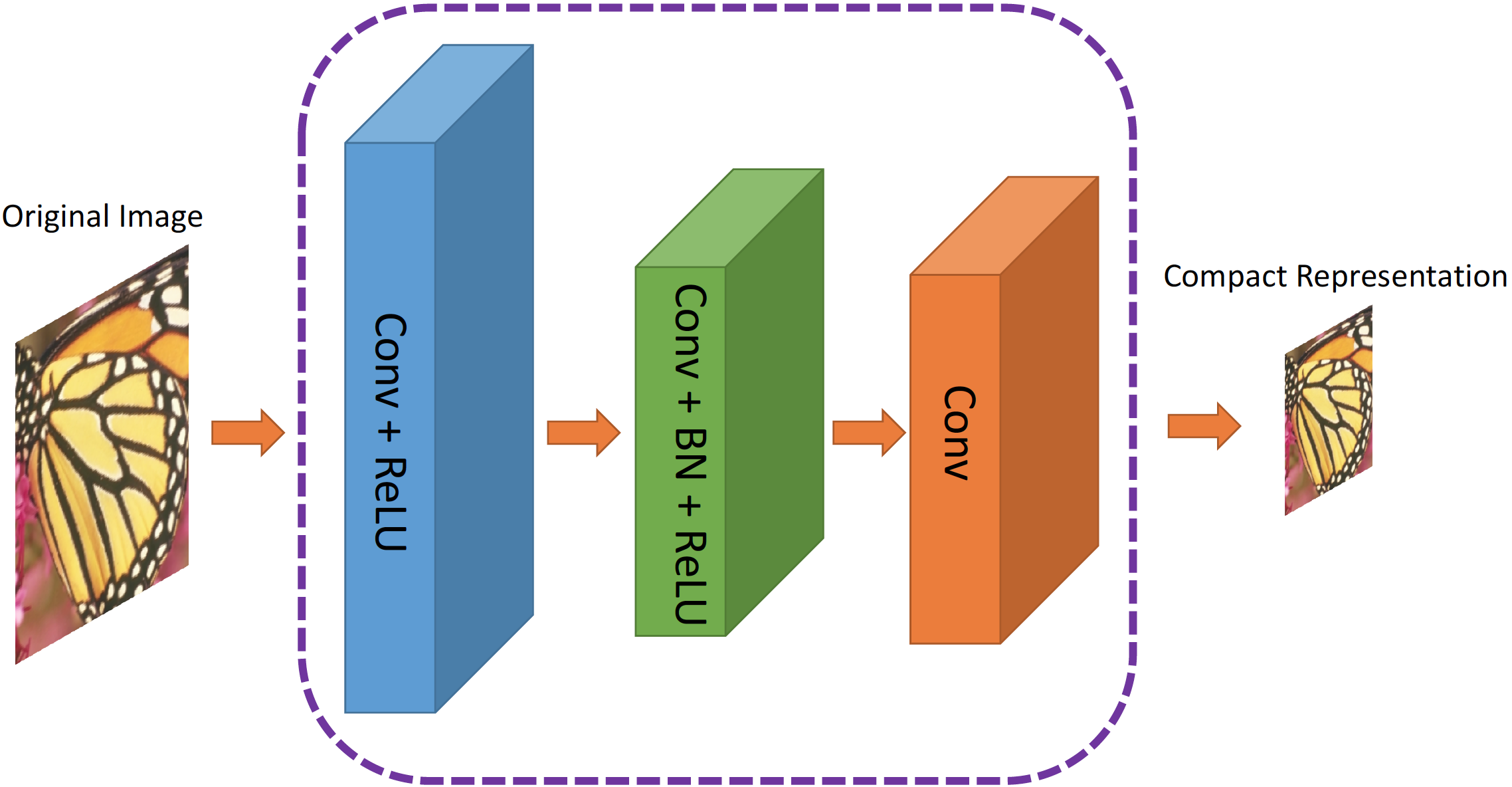}\\
\caption{ The architecture of the proposed ComCNN and feature maps of different layers.Upscaled image is obtained by Bicubic interpolation on Decoded image.}
\label{ComCNN}
\end{figure*}

\subsubsection{Reconstruction Convolutional Neural Network (RecCNN)}

As shown in Fig.\ref{RecCNN}, RecCNN is composed of 20 weight layers, which have three types of layer combinations: Convolution + ReLU, Convolution + Batch Normalization\cite{ioffe2015batch} + ReLU and Convolution. For the first layer, 64 filters of size $3\times 3\times c$ are used to generate 64 feature maps, followed by ReLU. For the 2nd to 19th layers, 64 filters of size $3\times 3\times 64$ are used, and batch normalization is added between convolution and ReLU. For the last layer, c filters of size $3\times 3\times 64$ are used to reconstruct the output. Residual learning and batch normalization are applied to speed up the training process and boost the performance. The compressed image is upsampled to the original image size using bicubic interpolation.

\begin{figure*}
\centering
\includegraphics[width=0.87\textwidth, height=6cm]{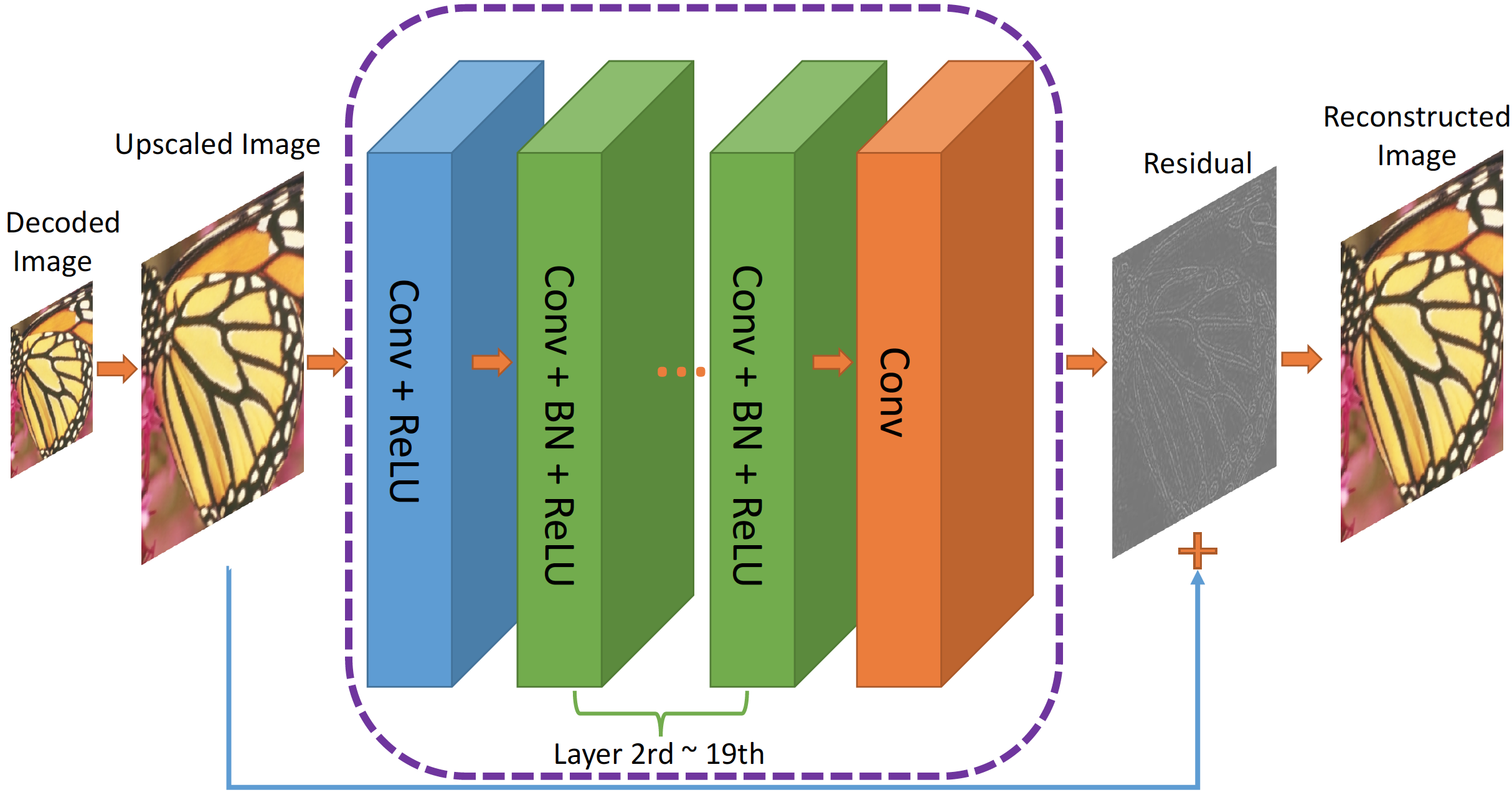}\\
\caption{ The architecture of the proposed RecCNN and feature maps of different layers. Upscaled image is obtained by bicubic interpolation on decoded image. The network predicts a residual image, then the sum of the input and the residual gives the high quality output.}
\label{RecCNN}
\end{figure*}

\subsection{Learning Algorithm}
According to the proposed architecture, both ComCNN and RecCNN try to make the reconstructed image as similar as possible to the original image. Therefore, the end-to-end optimization goal can be formulated as

\begin{eqnarray}
\left(\hat{\theta_1}, \hat{\theta_2}\right)=\arg\min\limits_{\theta_1,\theta_2}\norm{ Re\left(\theta_2, Co\left(Cr\left(\theta_1,x \right) \right) \right) - x }^2,
\end{eqnarray}
where $x$ represents the original image, $\theta_1$  and $\theta_2$ are the parameters of ComCNN and RecCNN, respectively. $Cr(\cdot)$  and $Re()$ represent the ComCNN and RecCNN, respectively. $Co(\cdot)$ represents an image codec (e.g. JPEG, JPEG2000 or BPG). From this objective function, we can see that an original image $x$ passes through the compression pipeline, including ComCNN, image codec and RecCNN, and finally outputs the reconstructed image $\hat{x}$. Such a process is an end-to-end compression.

Unfortunately, the $Co(\cdot)$ in Eq.(1) involves a rounding function, which is not differentiable when performing the back propagation algorithm. To solve this problem, we designed an iterative optimization learning algorithm. By fixing $\theta_2$, we can get
 \begin{eqnarray}
\hat{\theta_1}= \arg\min\limits_{\theta_1} \norm{ Re\left(\hat{\theta}_2, Co\left(Cr\left(\theta_1, x\right)\right)\right)-x }^2,
\end{eqnarray}
and by fixing $\theta_1$, we can obtain
\begin{eqnarray}
\hat{\theta_2}=\arg\min\limits_{\theta_2}\norm{ Re\left(\theta_2, Co\left(Cr\left(\hat{\theta}_1, x\right)\right)\right)-x }^2.
\end{eqnarray}

\subsubsection{Updating the Parameters $\theta_2$ of RecCNN}
 According to the network topology, an auxiliary variables $\hat{x}_m$  is introduced and defined as the decoded compact representation of $x$, which can be formulated as
\begin{eqnarray}
\hat{x}_m=Co\left(Cr\left(\hat{\theta}_1, x\right)\right).
\end{eqnarray}
After combining Eq.(4) and Eq.(3), we can obtain

\begin{eqnarray}
\hat{\theta}_2=\arg\min\limits_{\theta_2}\norm{ Re\left(\theta_2, \hat{x}_m \right) - x }^2.
\end{eqnarray}

\subsubsection{Updating the Parameters $\theta_1$ of ComCNN}
From Eq.(2), we can see that it is not a trivial task to obtain the optimal $\theta_1$ since the $Co(\cdot)$ is an inherently non-differentiable operation when performing back propagation. To solve this problem, we define an auxiliary variable $\hat{x}_m^*$ as the optimal input of RecCNN:
\begin{eqnarray}
\hat{x}_m^* =\arg\min\limits_{\hat{x}_m}\norm{ Re\left(\hat{\theta}_2, \hat{x}_m \right) - x }^2.
\end{eqnarray}

Here we make a reasonable and general assumption that $Re\left(\hat{\theta}_2, \cdot \right)$ is monotonic with respect to $\hat{x}_m^*$, which can be expressed as 

$\norm{\tau-\hat{x}_m^*}^2\geq \norm{\varphi-\hat{x}_m^*}^2 $, if and only if
\begin{eqnarray}
\norm{Re\left(\hat{\theta}_2, \tau \right)-x}^2 \geq \norm{Re\left(\hat{\theta}_2, \varphi \right)-x}^2.
\end{eqnarray}
Let $\tilde{\theta_1}$ be the solution of $\arg\min\limits_{\theta_1}\norm{Co\left(Cr\left(\theta, x\right)\right) - \hat{x}_m}^2$, i.e., for any possible $\theta_1^\prime$, it satisfies that
\begin{eqnarray}
\norm{Co\left(Cr\left(\theta_1^\prime, x\right)\right)-\hat{x}_m^*}^2 \geq \norm{Co\left(Cr\left(\tilde{\theta_1}, x\right)\right)-\hat{x}_m^*}^2.
\end{eqnarray}
Following assumption (7), we can obtain that
\begin{eqnarray}
\begin{split}
\norm{Re\left(\hat{\theta}_2, Co\left(Cr\left(\theta_1^\prime, x\right)\right)\right) - x}^2 \\
\geq \norm{Re\left(\hat{\theta}_2, Co\left(Cr\left(\tilde{\theta_1}, x\right)\right)\right) - x}^2.
\end{split}
\end{eqnarray}
Accordingly,
 \begin{eqnarray}
\tilde{\theta_1}= \arg\min\limits_{\theta_1} \norm{ Re\left(\hat{\theta}_2, Co\left(Cr\left(\theta_1, x\right)\right)\right)-x }^2.
\end{eqnarray}
Combining with Eq.(2), we can get $\hat{\theta_1} = \tilde{\theta_1}$, which is
\begin{eqnarray}
\hat{\theta}_1=\arg\min\limits_{\theta_1}\norm{ Co\left(Cr\left(\theta_1, x\right)\right)-\hat{x}_m^* }^2.
\end{eqnarray}
Since $Co(\cdot)$ is an codec, a reasonable solution of Eq.(12) is
\begin{eqnarray}
\hat{\theta}_1 \approx \arg\min\limits_{\theta_1}\norm{ Cr\left(\theta_1, x\right) -\hat{x}_m^* }^2.
\end{eqnarray}
Combine Eq.(13) and the assumption (7) above, it arrives 
\begin{eqnarray}
\hat{\theta}_1 = \arg\min\limits_{\theta_1}\norm{ Re\left(\hat{\theta}_2, Cr\left(\theta_1, x\right)\right) -x}^2.
\end{eqnarray}

Here, we get Eq.(13) with a reasonable assumption and rigorous derivations, which is the approximation of Eq.(2). In this paper, we use Eq.(13) to train ComCNN instead of Eq.(2).

We can obtain the optimal $\theta_1$ and $\theta_2$ by iteratively optimizing Eq.(5) and Eq.(13), respectively. In light of all derivations above, the complete description of the proposed algorithm is given in Algorithm \ref{algorithm1}.

\begin{algorithm}
\caption{The Proposed Compression Framework for Training Sub-Networks}
\label{algorithm1}
\begin{algorithmic}[1]
\State \textbf{Input:} The original image x
\State \textbf{Initialization:}Random initial $\hat{\theta}_1^0$ and $\hat{\theta}_2^0$
\For{$t=1 \to T$}
	\State Update $\hat{x}_m^t$ by computing Eq.(2)
	\For{$x_m = \hat{x}_m^t$}
		\State Update $\hat{\theta}_2^t$ by training RecCNN to compute Eq.(5)
	\EndFor
	\For{$\theta_2 = \hat{\theta}_2^t$}
		\State Update $\hat{\theta}_1^t$ by training ComCNN to compute
		\State Eq.(13)
	\EndFor
\EndFor
\State \textbf{Return:} $\hat{\theta}_1^t$, $\hat{\theta}_2^t$ and $\hat{x}_m^t$
\end{algorithmic}
\end{algorithm}

\subsection{Loss Functions}
\subsubsection{For ComCNN training}
Given a set of original images $x$ and trained parameters $\theta_2$, we use mean squared error (MSE) as the loss function
\begin{equation}
L_{1}(\theta_1)=\frac{1}{2N}\sum \limits_{k=1}^N \norm{ Re\left(\hat{\theta}_2, Cr\left(\theta_1, x_k\right)\right)-x_k }^2,
\end{equation}
where $N$ and $\theta_1$ represents the batch size and the trainable parameter, respectively.

\subsubsection{For RecCNN training}
Having obtained a set of compact representation $\hat{x}_m$ from ComCNN and original images $x$, we use MSE as the loss function:
\begin{equation}
L_{2}(\theta_2)=\frac{1}{2N}\sum \limits_{k=1}^N \norm{res\left(Co\left(\hat{x}_{m_k}\right),\theta_2 \right) -\left(Co\left( \hat{x}_{m_k} \right) - x_k \right)}^2,
\end{equation}
where $\theta_2$ represents the trainable parameter. $res(\cdot)$ represents the residual learned by RecCNN. Clearly, it looks somewhat different from Eq.(5), but they are not contradictory. Actually, they are essentially identical, and Eq.(15) is just expresses Eq.(5) as the form of the residual.

\section{Experiments}
To evaluate the performance of the proposed compression framework, we conduct experimental comparisons against standard compression methods (e.g., JPEG, JPEG 2000 and BPG) with a post-processing deblocking or denoising method. Five representative image deblocking methods, i.e. Sun'sš\cite{sun2007postprocessing}, DicTV\cite{chang2014reducing}, Zhang's\cite{zhang2013compression}, Ren'sš\cite{ren2013image}, Zhang's\cite{zhang2016concolor} and two representative image denoising methods, i.e. BM3D\cite{dabov2007image} and WNNM\cite{gu2014weighted} are chosen due to their state-of-the-art performance. Moreover, ARCNN\cite{dong2015compression} is also chosen since it is a landmark deblocking method  based on deep learning and achieves the state-of-the-art performance. Meanwhile, in order to demonstrate the effectiveness of ComCNN, we remove ComCNN in the framework and just using RecCNN to reconstruct the decoded image.  Similarly, we remove the RecCNN to examine the effect of ComCNN using bicubic interpolation to obtain the reconstructed image of the same size as the original image. The results of all the compared methods are obtained by running the source codes of the original authors  with the optimal parameters. Through this section, we use the name of the post-processing method to denote a compared method.

We use the MatConvNet package\cite{vedaldi2015matconvnet} to train the proposed networks. All experiments are carried out in the Matlab (R2015b) environment running on a computer with Inter(R) Xeon(R) CPU E5-2670 2.60GHz and an Nvidia Tesla K40c GPU. It takes about 28 hours and one day to train the ComCNN and RecCNN on GPU, respectively.

\subsection{Datasets for Training and Testing}

Following\cite{chen2015trainable}, we use 400 images of size $180 \times 180$ for training. A total of 204800 ($400 \times 8 \times 64$)\footnote{ For each image, there are 8 augmentations and 64 patches of size $40 \times 40 $ extracted.} patches are sampled with a stride of 20 on the training images as well as their augmentations (flip and rotate with different angles). Our experiments indicate that using a larger training set can only bring negligible performance improvements. For testing, we use 7 images as shown in Fig.\ref{TestSet} , which are widely used in the literature. Please note that all the test images are not included in the training dataset.

\begin{figure*}
\centering
\includegraphics[width=1.0\textwidth]{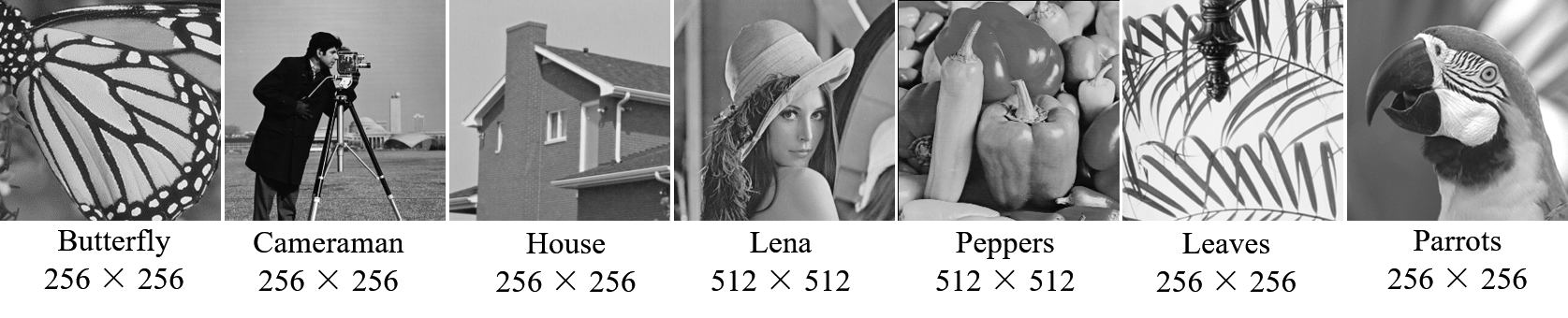}\\
\caption{ The used test images.}
\label{TestSet}
\end{figure*}

\subsection{Model Initialization}
We initialize the weights of ComCNN using the method in \cite{he2015delving} and use Adam  algorithm \cite{kingma2014adam} with $\alpha=0.001, \beta_1=0.9, \beta_2=0.999 $ and $\varepsilon = 10^{-8}$. We train ComCNN for 50 epochs using a batch size of 128. The learning rate is decayed exponentially from 0.01 to 0.0001 for 50 epochs. The weights initialization and gradient updating of RecCNN is the same as ComCNN. RecCNN is also trained for 50 epochs using the same batch size with ComCNN. The learning rate is decayed exponentially from 0.1 to 0.0001 for 50 epochs.

\subsection{Experimental Results}
In our experiments, we set different quality factors (QF) to achieve the same bits per pixel (bpp) for both the proposed and compared methods. For the proposed method, we first manually adjust the QF for the compression of compact representation by JPEG to achieve almost the same bpp with the compared image enhancement methods. Then we compare the PSNR and SSIM of the proposed method with the compared methods. The comparison results for all test images with QF = 5 and QF = 10 are provided in Table \ref{qf=5} and Table \ref{qf=10}, respectively, where the best results are highlighted in bold.

\begin{table*}[!t]
\centering
\caption{JPEG: QF = 5, PSNR (dB) and SSIM results of all competitive algorithms grayscale image deblocking and denoising.}
\label{qf=5}
\renewcommand\arraystretch{1.4}
\begin{small}
\begin{tabular}{>{\hfil}p{80pt}<{\hfil}||>{\hfil}p{35pt}<{\hfil}>{\hfil}p{55pt}<{\hfil}>{\hfil}p{35pt}<{\hfil}>{\hfil}p{35pt}<{\hfil}>{\hfil}p{35pt}<{\hfil}>{\hfil}p{35pt}<{\hfil}>{\hfil}p{35pt}<{\hfil}||>{\hfil}p{35pt}<{\hfil}}
\hline\hline
{\bfseries Test Images} & {\bfseries Butterfly} & {\bfseries Carmerman} & {\bfseries House}  & {\bfseries Lena} & {\bfseries Peppers} & {\bfseries Leaves} & {\bfseries Parrots} & {\bfseries Average} \\
\hline\hline

\multicolumn{9}{c}{PSNR} \\
\hline\hline

 JPEG & 22.58	& 24.45 & 27.77 & 27.33 & 27.17 &	22.49 & 26.19 & 25.43 \\

 Sun's\cite{sun2007postprocessing} & 23.83 & 25.25 & 29.09 & 28.87 & 29.05 & 23.47 & 27.45 & 26.72 \\

Zhang's\cite{zhang2013compression} & 24.20 & 25.39 & 29.24 & 29.00 & 29.07	& 24.13 & 27.78 & 26.97 \\

Ren's\cite{ren2013image} & 24.58 & 25.46 & 29.66 & 29.07 & 29.07 & 24.56 & 27.87 & 27.18 \\

BM3D\cite{dabov2007image} & 24.05 & 25.27 & 29.21 & 28.63 &	28.52 & 24.02 & 27.33 & 26.72 \\

DicTV\cite{chang2014reducing} & 23.10 & 24.54 & 28.45 & 28.07 &	27.95 & 23.01 & 26.83 & 25.99 \\

WNNM\cite{gu2014weighted} & 24.75 & 25.49 & 29.62 & 28.95 &	28.99 & 24.68 & 27.80 & 27.18 \\

Zhang's\cite{zhang2016concolor} & 25.30 & 25.61 & 30.12 & 29.51 & 29.61 & 24.99 & 28.27 & 27.63 \\

ARCNN\cite{dong2015compression} & 25.64 & 25.27 & 29.68 & 29.31 & 29.02 & 25.07 & 28.13 & 27.45 \\

ComCNN & 23.05 & 24.93 & 29.17 & 29.46 & 29.33 & 23.19 & 27.67 & 26.69\\

RecCNN & 25.99 & 26.33 & 30.13 & 29.63 & 29.81 & 25.21 & 28.52 & 28.02 \\
\hline

{\bfseries Proposed} & {\bfseries 26.23} & {\bfseries 26.53} & {\bfseries 31.45} & {\bfseries 31.14} & {\bfseries 30.84} & {\bfseries 25.52} & {\bfseries 30.12} & {\bfseries 28.83} \\
\hline\hline

\multicolumn{9}{c}{SSIM} \\
\hline\hline

 JPEG & 0.7378	& 0.7283 & 0.7733 & 0.7367 & 0.7087 & 0.7775 & 0.7581 & 0.7456 \\

 Sun's\cite{sun2007postprocessing} & 0.8321 & 0.7687 & 0.8113 & 0.8061 & 0.7931 & 0.8380 & 0.8323 & 0.8104 \\

Zhang's\cite{zhang2013compression} & 0.8313 & 0.7672 & 0.8141 & 0.8035 & 0.7895 & 0.8548 & 0.8308 & 0.8130 \\

Ren's\cite{ren2013image} & 0.8419 & 0.7666 & 0.8197 & 0.8010 & 0.7876 & 0.8720 & 0.8310 & 0.8171\\

BM3D\cite{dabov2007image} & 0.8184 & 0.7607 & 0.8082 & 0.7837 & 0.7639 & 0.8510 & 0.8118 & 0.7997\\

DicTV\cite{chang2014reducing} & 0.7769 & 0.6658 & 0.7963 & 0.7744 & 0.7456 & 0.8104 & 0.8005 & 0.7671\\

WNNM\cite{gu2014weighted} & 0.8445 & 0.7674 & 0.8178 & 0.7947 & 0.7827 & 0.8749 & 0.8287 & 0.8158\\

Zhang's\cite{zhang2016concolor} & 0.8667 & 0.7666 & 0.8285 & 0.8169 & 0.8031 & 0.8882 & 0.8460 & 0.8308 \\

ARCNN\cite{dong2015compression} & 0.8741 & 0.7674 & 0.8209 & 0.8142 & 0.7961 & 0.8983 & 0.8446 & 0.8308 \\

ComCNN & 0.7488 & 0.7662 & 0.8119 & 0.8042  & 0.7966 & 0.7885 & 0.8377 & 0.7934\\

RecCNN & 0.8760 & 0.7945 & 0.8251 & 0.8195 & 0.8004 & 0.8803 & 0.8497 & 0.8351 \\
\hline

{\bfseries Proposed} & {\bfseries 0.8847 } & {\bfseries 0.8167 } & {\bfseries 0.8456 } & {\bfseries 0.8486 } & {\bfseries 0.8328 } & {\bfseries 0.8912 } & {\bfseries 0.8951 } & {\bfseries 0.8535 }\\
\hline\hline

\end{tabular}
\end{small}
\end{table*}

\begin{table*}[!t]
\centering
\caption{JPEG: QF = 10, PSNR (dB) and SSIM results of all competitive algorithms grayscale image deblocking and denoising.}
\label{qf=10}
\renewcommand\arraystretch{1.4}
\begin{small}
\begin{tabular}{>{\hfil}p{80pt}<{\hfil}||>{\hfil}p{35pt}<{\hfil}>{\hfil}p{55pt}<{\hfil}>{\hfil}p{35pt}<{\hfil}>{\hfil}p{35pt}<{\hfil}>{\hfil}p{35pt}<{\hfil}>{\hfil}p{35pt}<{\hfil}>{\hfil}p{35pt}<{\hfil}||>{\hfil}p{35pt}<{\hfil}}
\hline\hline
{\bfseries Test Images} & {\bfseries Butterfly} & {\bfseries Carmerman} & {\bfseries House}  & {\bfseries Lena} & {\bfseries Peppers} & {\bfseries Leaves} & {\bfseries Parrots} & {\bfseries Average} \\
\hline\hline

\multicolumn{9}{c}{PSNR} \\
\hline\hline

JPEG & 25.24 & 26.47 & 30.56 & 30.41 & 30.14 & 25.40 & 28.96 & 28.17\\

Sun's\cite{sun2007postprocessing} & 26.52 & 27.26 & 32.00 & 31.72 & 31.62 & 26.60 & 30.04 & 29.39\\

Zhang's\cite{zhang2013compression} & 26.83 & 27.45 & 32.11 & 31.92 & 31.68 & 27.26 & 30.50 & 29.68\\

Ren's\cite{ren2013image} & 27.17 & 27.43 & 32.41 & 31.92 & 31.63 & 27.59 & 30.34 & 29.78\\

BM3D\cite{dabov2007image} & 26.64 & 27.25 & 32.07 & 31.77 & 31.42 & 26.98 & 30.05 & 29.45\\

DicTV\cite{chang2014reducing} & 26.09 & 26.92 & 31.77 & 31.55 & 31.29 & 26.33 & 29.82 & 29.11\\

WNNM\cite{gu2014weighted} & 27.22 & 27.40 & 32.42 & 31.93 & 31.64 & 27.66 & 30.33 & 29.80\\

Zhang's\cite{zhang2016concolor} & 27.09 & {\bfseries 27.72} & 33.04 & 32.19 & {\bfseries 31.94} & 28.20 & 30.66 & 30.12\\

ARCNN\cite{dong2015compression} & 28.54 & 27.62 & 32.53 & 32.05 & 31.50 & 28.31 & 30.62 & 30.16\\

ComCNN & 26.32 & 27.05 & 31.82 & 31.63 & 31.04 & 26.51 & 29.97 & 29.12 \\

RecCNN & 28.04 & 27.33 & 32.76 & 32.35 & 31.34 & 28.53 & 30.85 & 30.17\\
\hline

{\bfseries Proposed} & {\bfseries 28.60 } & 27.44 & {\bfseries 33.25 } & {\bfseries 33.11 } & 31.83 & {\bfseries 28.77 } & {\bfseries 31.11 } & {\bfseries 30.59 }\\
\hline\hline

\multicolumn{9}{c}{SSIM} \\
\hline\hline

 JPEG & 0.8325 & 0.7965 & 0.8183 & 0.8183 & 0.7839 & 0.8609 & 0.8336 & 0.8206 \\

Sun's\cite{sun2007postprocessing} & 0.8871 & 0.8358 & 0.8504 & 0.8590 & 0.8322 & 0.9138 & 0.8783 & 0.8652\\

Zhang's\cite{zhang2013compression} & 0.8923 & 0.8329 & 0.8513 & 0.8597 & 0.8317 & 0.9212 & 0.8804 & 0.8671\\

Ren's\cite{ren2013image} & 0.9010 & 0.8259 & 0.8526 & 0.8571 & 0.8300 & 0.9309 & 0.8775 & 0.8679\\

BM3D\cite{dabov2007image} & 0.8896 & 0.8240 & 0.8492 & 0.8549 & 0.8250 & 0.9207 & 0.8749 & 0.8626\\

DicTV\cite{chang2014reducing} & 0.8699 & 0.8046 & 0.8484 & 0.8559 & 0.8244 & 0.9032 & 0.8741 & 0.8544\\

WNNM\cite{gu2014weighted} & 0.9019 & 0.8248 & 0.8531 & 0.8571 & 0.8303 & 0.9325 & 0.8775 & 0.8681\\

Zhang's\cite{zhang2016concolor} & 0.9142 & 0.8401 & 0.8609 & 0.8661 & 0.8358 & 0.9406 & 0.8842 & 0.8774 \\

ARCNN\cite{dong2015compression} & 0.9237 & 0.8389 & 0.8591 & 0.8711 & 0.8434 & {\bfseries 0.9495} & 0.8942 & 0.8828\\

ComCNN & 0.8796 & 0.8154 & 0.8497 & 0.8573 & 0.8296 &  0.9095 & 0.8766 & 0.8597 \\

RecCNN & 0.9192 & 0.8429 & 0.8584 & 0.8679 & 0.8396 & 0.9443 & 0.8884 & 0.8801\\
\hline

{\bfseries Proposed} & {\bfseries 0.9245} & {\bfseries 0.8448} & {\bfseries 0.8678} & {\bfseries 0.8838} & {\bfseries 0.8532} & 0.9475 & {\bfseries 0.9047} & {\bfseries 0.8895} \\
\hline\hline

\end{tabular}
\end{small}
\end{table*}
As seen from Table \ref{qf=5} and Table \ref{qf=10}, in the case of QF = 5, the proposed compression framework achieves 1.20dB gains in PSNR and 0.0227 gains in SSIM compared against Zhang's \cite{zhang2016concolor}, which is state-of-the-art in the compared methods. It is worth mentioning that the proposed framework outperforms all the compared image enhancement methods including ARCNN\cite{dong2015compression}, which is a milestone based on CNN. Meanwhile, in the case of QF = 10, the proposed compression framework achieves 0.43dB and 0.0067 gains in PSNR and SSIM, respectively, compared against ARCNN\cite{dong2015compression}. The visual quality comparisons in the case of QF = 5 for \textit{Lena} is provided in Fig.~\ref{Lena512}. We can see that the blocking artifacts are obvious in the image decoded directly by the standard JPEG. DicTV\cite{chang2014reducing}, Sun's\cite{sun2007postprocessing}, WNNM\cite{gu2014weighted} and BM3D\cite{dabov2007image} remove the artifacts partially, but there are still some artifacts visible in the reconstructed image. Zhang's~\cite{zhang2013compression} and Ren's~\cite{ren2013image} generate better results than Sun's\cite{sun2007postprocessing} and BM3D\cite{dabov2007image}. However, the blur effects along the edges are generated at the same time. Zhang's~\cite{zhang2016concolor} achieves a better PSNR and SSIM, but it makes the image over-smoothing and discards some details in image edges. ARCNN\cite{dong2015compression} and ReCNN achieve better visual quality than other compared methods. The proposed compression framework not only removes most of the artifacts significantly, but also preserves more details on both edges and textures than all the compared methods. In order to verify the effect of ComCNN, we remove ComCNN and use RecCNN alone to reconstruct the decoded image. Similarly, we remove RecCNN and only use ComCNN and bicubic interpolation to examine the effect of RecCNN. As shown in Table \ref{qf=5} and Table \ref{qf=10}, worse performances are obtained only with ComCNN or RecCNN. In addition, we show examples of the compact representation produced by ComCNN in Fig.\ref{com_re}. It can be seen that the compact representation maintains the structural information of the original image, but it is different from traditional down sampling methods. In a nutshell, both ComCNN and RecCNN play key roles in the proposed compression framework. Due to the collaboration of ComCNN and RecCNN, the compact representation preserves more useful information for the final image reconstruction. Our testing codes  are available in GitHub \footnote{\url{https://github.com/compression-framework/compression_framwork_for_tesing}}.

\begin{figure*}
\centering
\includegraphics[width=1.0\textwidth]{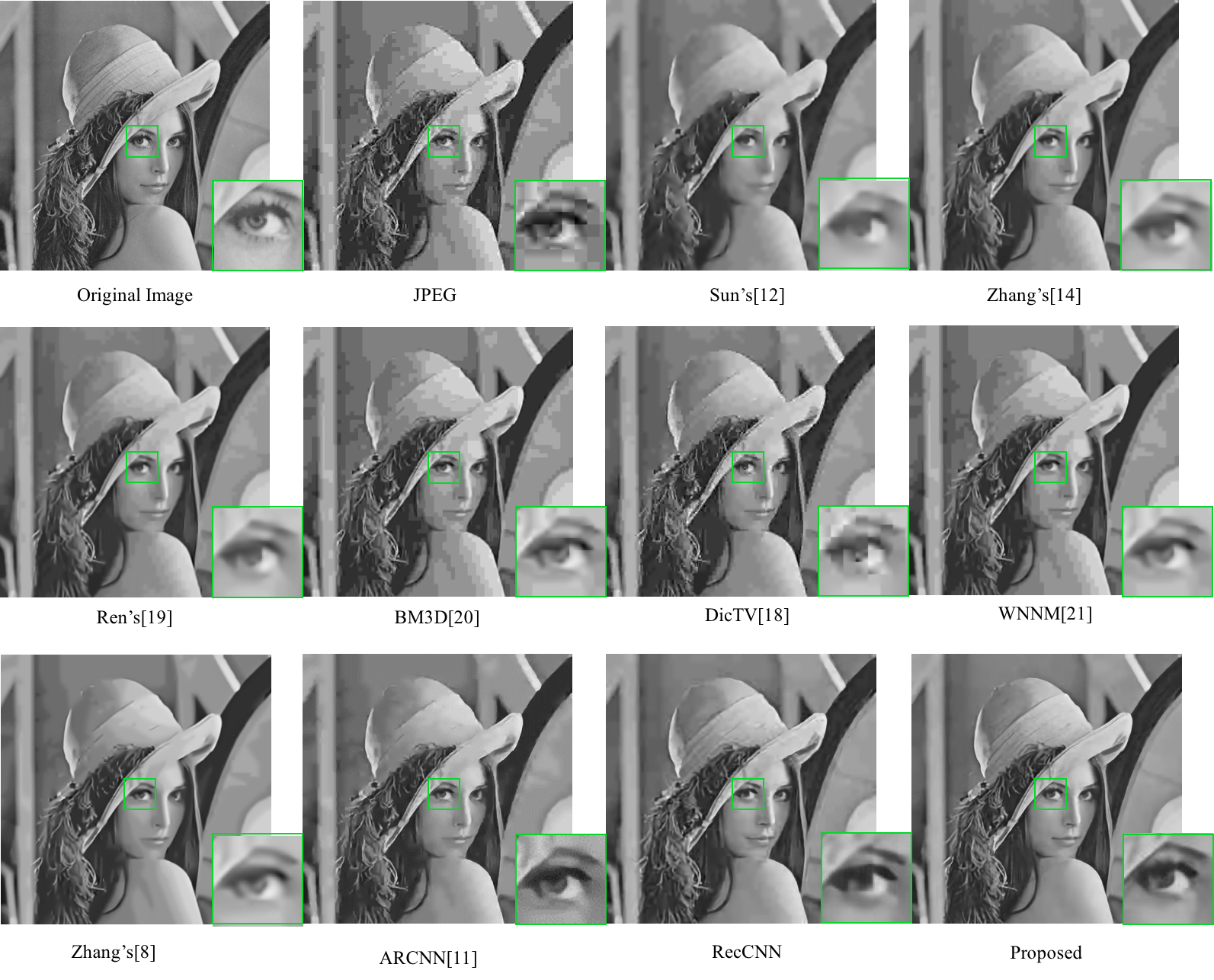}\\
\caption{ Visual quality comparison of image deblocking for \textit{Lena} in the case of QF = 5. From left to right and top to bottom: original image, JPEG(PSNR = 27.33 dB, SSIM = 0.7367), the deblocking results by Sun's (PSNR = 28.87 dB, SSIM = 0.8061), Zhang's (PSNR = 29.00 dB, SSIM = 0.8035), Ren's (PSNR = 29.07 dB, SSIM = 0.8010), BM3D (PSNR = 28.63 dB, SSIM = 0.7837), DicTV (PSNR = 28.07 dB, SSIM = 0.7744), WNNM (PSNR = 28.95 dB, SSIM = 0.7947), Zhang's (PSNR = 29.31 dB, SSIM = 0.8169), ARCNN (PSNR = 29.31 dB, SSIM = 0.8142), RecCNN (PSNR = 29.63 dB, SSIM = 0.8195) and the proposed framework (PSNR = 31.14 dB, SSIM = 0.8486)}.
\label{Lena512}
\end{figure*}

\begin{figure}
\centering
\includegraphics[width=8cm, height=8cm]{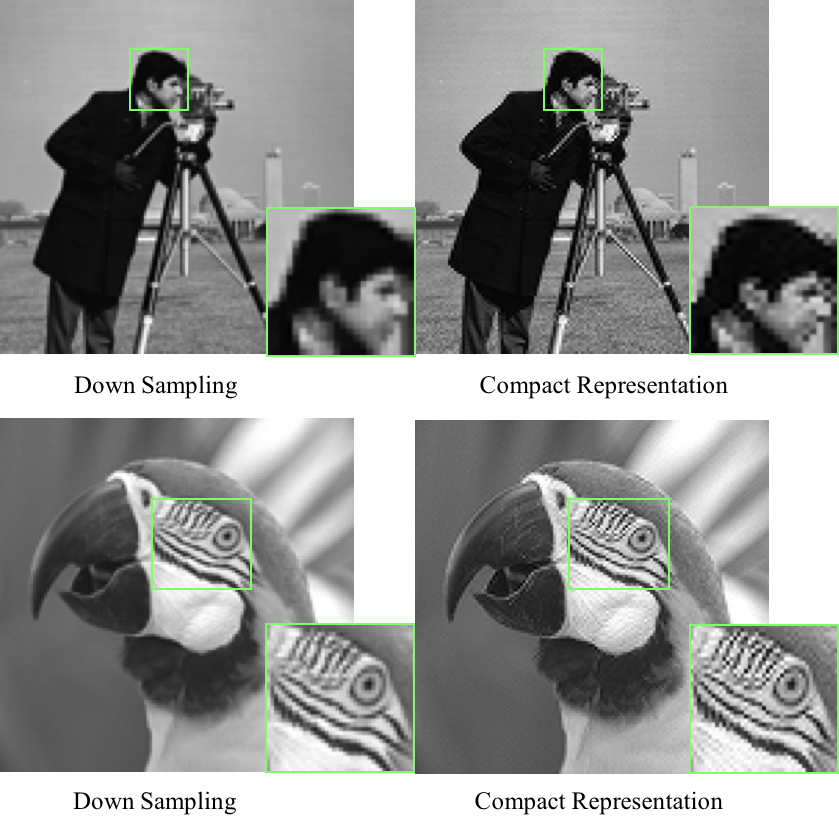}\\
\caption{ Visual quality comparison of the down sampling using bicubic interpolation and the compact representation produced by ComCNN. }
\label{com_re}
\end{figure}

We also evaluate our framework on JPEG 2000 and BPG (Better Portable Graphics)\footnote{F. Bellard, The BPG Image Format, http://bellard.org/bpg/} and achieve excellent performance. BPG compression is based on the High Efficiency Video Coding (HEVC), which is considered as a major advance in compression techniques. For JPEG2000, we test the proposed compression framework at different bit rates (from 0.1bpp to 0.4bpp) and compare it with JPEG 2000. Table \ref{jpeg2000} presents the comparison results with JPEG 2000. It can seen that our framework significantly outperforms JPEG2000 on all test bit-rates of all test images in terms of both PSNR and SSIM. For bpp from 0.1 to 0.4, the proposed framework achieves on average 3.06dB, 2.45dB, 1.34dB, 1.09dB and 0.1047, 0.0709, 0.0525, 0.0435 gains in PSNR and SSIM compared against JPEG2000. In Fig.~\ref{Parrots}, one can see that the proposed compression framework achieves much better subjective performance than JPEG2000, especially at very low bit rate. Our framework preserves more high-frequency information and recovers sharp edges and pure textures in the reconstructed image. For BPG, we test the BPG codec at QP (quality parameter) = 43 and 47. Further, we keep the bit-rates of the proposed compression framework almost the same for each image. The results are shown in Table \ref{bpg_table}. One can see that, if we treat RecCNN as a post-processing method, RecCNN achieves on average 0.81dB and 0.0168 gains in PSNR and SSIM. And the proposed compression framework achieves on average 0.99dB and 0.0218 gains in PSNR and SSIM while saving 5.22\% bit-rates. It is worth noting that the performance of our proposed compression framework on BPG is not so obvious on JPEG and JPEG2000 when compared, because BPG is already a very good compression method, which might not be significantly improved further.

\begin{figure*}
\centering
\includegraphics[width=1.0\textwidth]{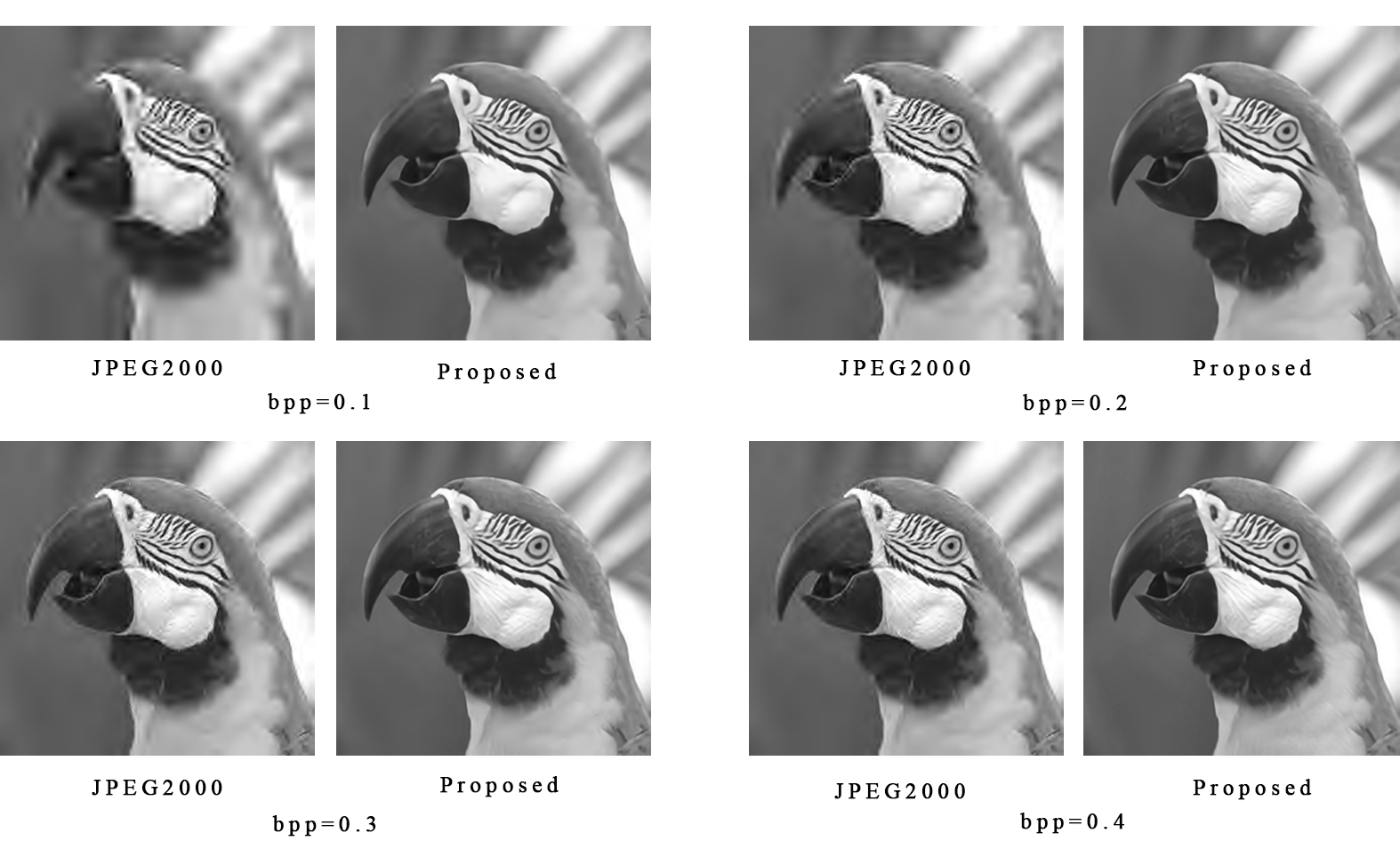}\\
\caption{ Subjective performance comparison of JPEG2000 when the bit rate from 0.1bpp to 0.4bpp for \textit{Parrot}. From left to right and top to bottom, the corresponding PSNR(in dB) and SSIM values are (26.05, 0.7853), (31.22, 0.8895), (29.96, 0.8589), (32.72, 0.9242), (32.42, 0.8897), (33.48, 0.9382), (34.42, 0.9150) and (35.09, 0.9480).}
\label{Parrots}
\end{figure*}

\begin{table*}[!t]
\centering
\caption{JPEG2000: PSNR (dB) and SSIM results of all competitive algorithms grayscale image deblocking and denoising.}
\label{jpeg2000}
\renewcommand\arraystretch{1.4}
\begin{small}
\begin{tabular}{c||cc|cc|cc|cc}
\hline\hline
\multirow{2}{100pt}{\diagbox{Test Images}{Rate(bpp)}}& \multicolumn{2}{c|}{0.1} & \multicolumn{2}{c|}{0.2} & \multicolumn{2}{c|}{0.3} & \multicolumn{2}{c}{0.4} \\
\cline{2-9}
 & JPEG2000 & Proposed & JPEG2000 & Proposed & JPEG2000 & Proposed & JPEG2000 & Proposed
 \\
\hline\hline

\multicolumn{9}{c}{PSNR} \\
\hline\hline

Butterfly & 18.92 & {\bfseries 23.66 } & 21.99 & {\bfseries 26.73 } & 24.01 & {\bfseries 27.47 } & 25.44 & {\bfseries 29.43 }\\

Cameraman & 23.33 & {\bfseries 26.66 } & 26.23 & {\bfseries 28.44 } & 28.30 & {\bfseries 29.35 } & 29.80 & {\bfseries 30.41 }\\

House & 28.02 & {\bfseries 30.54 } & 31.92 & {\bfseries 34.34 } & 34.04 & {\bfseries 35.30 } & 35.35 & {\bfseries 35.76 }\\

Lena & 29.88 & {\bfseries 31.07 } & 32.98 & {\bfseries 34.68 } & 34.81 & {\bfseries 35.65 } & 36.13 & {\bfseries 36.53 }\\

Parrots & 26.05 & {\bfseries 31.22 } & 29.96 & {\bfseries 32.72 } & 32.42 & {\bfseries 33.48 } & 34.42 & {\bfseries 35.09 }\\


Peppers & 29.66 & {\bfseries 31.07 } & 32.56 & {\bfseries 33.43 } & 34.11 & {\bfseries 34.49 } & 35.02 & {\bfseries 35.46 }\\
\hline

{\bfseries Average} & 25.98 & {\bfseries 27.37 } & 29.27 & {\bfseries 31.72 } & 31.28 & {\bfseries 32.62 } & 32.69 & {\bfseries 33.78 } \\
\hline\hline

\multicolumn{9}{c}{SSIM} \\
\hline\hline
 Butterfly & 0.5888 & {\bfseries 0.8308 } & 0.7257 & {\bfseries 0.8866 } & 0.7968 & {\bfseries 0.8894 } & 0.8419 & {\bfseries 0.9266 }\\

Cameraman & 0.6764 & {\bfseries 0.8238 } & 0.7659 & {\bfseries 0.8667 } & 0.8146 & {\bfseries 0.8955 } & 0.8482 & {\bfseries 0.9026 }\\

House & 0.7702 & {\bfseries 0.8310 } & 0.8375 & {\bfseries 0.8777 } & 0.8662 & {\bfseries 0.8989 } & 0.8823 & {\bfseries 0.9122 }\\

Lena & 0.8176 & {\bfseries 0.8492 } & 0.8763 & {\bfseries 0.9002 } & 0.8973 & {\bfseries 0.9183 } & 0.9143 & {\bfseries 0.9320 }\\

Parrots & 0.7853 & {\bfseries 0.8895 } & 0.8589 & {\bfseries 0.9242 } & 0.8897 & {\bfseries 0.9382 } & 0.9150 & {\bfseries 0.9480 }\\


Peppers & 0.7851 & {\bfseries 0.8273 } & 0.8368 & {\bfseries 0.8733 } & 0.8591 & {\bfseries 0.8988 } & 0.8719 & {\bfseries 0.9127 }\\
\hline

{\bfseries Average} & 0.7273 & {\bfseries 0.8419 } & 0.8169 & {\bfseries 0.8878 } & 0.8540 & {\bfseries 0.9065 } & 0.8789 & {\bfseries 0.9224 } \\
\hline\hline

\end{tabular}
\end{small}
\end{table*}

\begin{table*}[!t]
\centering
\caption{BPG: PSNR (dB) and SSIM Results of BPG, BPG + RecCNN and the Proposed Method  .}
\label{bpg_table}
\renewcommand\arraystretch{1.4}
\renewcommand{\multirowsetup}{\centering}
\begin{small}
\begin{tabular}{c||c|c|c|c|c|c|c}
\hline\hline
\multirow{2}{50pt}{Test Images}& \multicolumn{2}{c|}{BPG} & \multicolumn{2}{c|}{BPG + RecCNN} & \multicolumn{3}{c}{ComCNN + BPG + RecCNN} \\
\cline{2-8}
 & Size(Bytes) & PSNR/SSIM & Size(Bytes) & PSNR/SSIM & Size(Bytes) & PSNR/SSIM & Bits Saving \\
\hline

\multirow{2}{50pt}{Cameraman} & 752 & 25.95/0.7723 & 752 & 26.55/0.7867 & 642 & \bf{26.70/0.7887} & 14.63\% \\
\cline{2-8}
 & 1251 & 28.12/0.8188 & 1251 & 28.77/0.8288 & 1105 & \bf{28.85/0.8325} & 11.67\% \\
\hline

\multirow{2}{50pt}{House} & 389 & 29.06/0.8023 & 389 & 29.61/0.8121 & 384 & \bf{30.21/0.8274} & 1.29\% \\
\cline{2-8}
 & 592 & 31.35/0.8349 & 592 & 32.02/0.8433 & 524 & \bf{32.31/0.8489} & 11.49\% \\
\hline

\multirow{2}{50pt}{Lena} & 1616 & 28.79/0.7896 & 1616 & 29.20/0.8016 & 1585 & \bf{29.47/0.8066} & 1.92\% \\
\cline{2-8}
 & 2737 & 30.97/0.8351 & 2737 & 31.46/0.8456 & 2702 & \bf{31.74/0.8529} & 1.28\% \\
\hline

\multirow{2}{50pt}{Butterfly} & 1277 & 24.43/0.8335 & 1277 & \bf{25.79/0.8691} & 1203 & 25.65/0.8648 & 5.79\% \\
\cline{2-8}
 & 1999 & 26.86/0.8875 & 1999 & 28.03/0.9069 &  1934 & \bf{28.09/0.9076} & 3.25\% \\
\hline

\multirow{2}{50pt}{Peppers} & 1797 & 28.93/0.7759 & 1797 & 29.59/0.7944 & 1722 & \bf{29.75/0.7978} & 4.17\% \\
\cline{2-8}
 & 2747 & 30.81/0.8108 & 2747 & 31.38/0.8220 & 2699 & \bf{31.43/0.8287} & 1.75\% \\
\hline

\multirow{2}{50pt}{Leaves} & 1651 & 24.55/0.8670 & 1651 & 25.88/0.9046 & 1564 & \bf{25.97/0.9090} & 5.27\% \\
\cline{2-8}
 & 2492 & 27.17/0.9195 & 2492 & 28.71/0.9405 & 2423 & \bf{28.83/0.9476} & 2.77\% \\
\hline

\multirow{2}{50pt}{Parrots} & 621 & 27.94/0.8235 & 621 & 28.61/0.8389 & 595 & \bf{28.85/0.8483} & 4.19\% \\
\cline{2-8}
 & 1017 & 30.25/0.8562 & 1017 & 30.95/0.8680 & 980 & \bf{31.19/0.8716} & 3.64\% \\
\hline
\bf{Average} & - & 28.23/0.8305 & - & 29.04/0.8473 & - & \bf{29.22/0.8523} & 5.22\% \\
\hline\hline

\end{tabular}
\end{small}
\end{table*}

In order to show the effectiveness of the proposed compression framework, we test our method on Set5\cite{kim2015accurate}, Set14\cite{kim2015accurate}, LIVE1\cite{sheikh2005live} and General-100\cite{dong2016accelerating} datasets. It is worth mentioning that the General-100 dataset contains 100 bmp-format images  with no compression, which are very suitable for compression task. Results are shown  in Table \ref{dataset}, from which we can see that the performance of our proposed compression exceeds JPEG and JPEG2000 by a larger margin for all four testing datasets.

\subsection{Running Time}
The running time of all compared methods when dealing with a $256 \times 256$ grayscale image in CPU or GPU are shown in Table \ref{RunningTime}. It should be noted that it is not possible to test the running time in GPU for all other compared methods. As we can see from Table~\ref{RunningTime}, the proposed framework needs only 1.56s and 0.017s in CPU and GPU, respectively. Our compression framework is faster than other post-processing methods. In addition, we also calculate the running time of our sub-network RecCNN, which almost takes the entire running time of our method. Because RecCNN has 20 layers, which is much deeper than ComCNN with only 3 layers.

\begin{table*}[!t]
\centering
\caption{ Running time (s) of compared methods in CPU (/GPU) tested on a $256 \times 256$ grayscale image.}
\label{RunningTime}
\renewcommand\arraystretch{1.3}
\begin{small}
\begin{tabular}{>{\hfil}p{40pt}<{\hfil}|>{\hfil}p{48pt}<{\hfil}|>{\hfil}p{40pt}<{\hfil}|>{\hfil}p{43pt}<{\hfil}|>{\hfil}p{43pt}<{\hfil}|>{\hfil}p{48pt}<{\hfil}|>{\hfil}p{45pt}<{\hfil}|>{\hfil}p{50pt}<{\hfil}|>{\hfil}p{45pt}<{\hfil}}
\hline\hline

Sun's\cite{sun2007postprocessing} & Zhang's\cite{zhang2013compression} & Ren's\cite{ren2013image}           & BM3D\cite{dabov2007image} & DicTV\cite{chang2014reducing} & WNNM\cite{gu2014weighted} & Zhang's\cite{zhang2016concolor} & RecCNN & Proposed\\
\hline
132/- & 251/- & 36/- & 3.40/- & 53/- & 230/- & 442/- & 1.50/0.015  & 1.56/0.017\\
\hline\hline

\end{tabular}
\end{small}
\end{table*}

\begin{table}[!t]
\centering
\caption{Average PSNR(dB)/SSIM results of JPEG and the proposed method for quality factors 5 and 10, JPEG2000 and the proposed method for bpp 0.1, 0.2, 0.3 for Set5, Set14, LIVE1 and General-100 }
\label{dataset}
\renewcommand\arraystretch{1.4}
\begin{small}
\begin{tabular}{>{\hfil}p{30pt}<{\hfil}||>{\hfil}p{30pt}<{\hfil}|>{\hfil}p{65pt}<{\hfil}|>{\hfil}p{65pt}<{\hfil}}
\hline\hline
\multicolumn{4}{c}{\bf{JPEG}} \\
\hline\hline

\multirow{2}{30pt}{DataSets} & Quality & JPEG & Proposed \\
\cline{3-4}
 &  Factors & PSNR/SSIM & PSNR/SSIM \\
\hline

\multirow{2}{30pt}{Set5} &  5 & 26.13/0.7206 & {\bfseries 29.20/0.8387} \\
\cline{2-4}
& 10 & 28.99/0.8109 & {\bfseries 31.40/0.8854} \\
\hline

\multirow{2}{30pt}{Set14} &  5 & 24.90/0.6686 & {\bfseries 26.89/0.7914} \\
\cline{2-4}
& 10 & 27.49/0.7762 & {\bfseries 28.91/0.8336} \\
\hline

\multirow{2}{30pt}{LIVE1} &  5 & 24.60/0.6666 & {\bfseries 26.78/0.7934} \\
\cline{2-4}
& 10 & 27.02/0.7720 & {\bfseries 28.84/0.8411} \\
\hline

\multirow{2}{30pt}{Genreal-100} &  5 & 25.93/0.7228 & {\bfseries 27.29/0.8447} \\
\cline{2-4}
& 10 & 28.92/0.8199 & {\bfseries 30.16/0.8767} \\
\hline\hline

\multicolumn{4}{c}{\bf{JPEG2000}} \\
\hline\hline

\multirow{2}{30pt}{DataSets} & \multirow{2}{28pt}{Rate(bpp)} & JPEG2000 & Proposed \\
\cline{3-4}
 & & PSNR/SSIM & PSNR/SSIM \\
\hline

\multirow{2}{30pt}{Set5} &  0.1 & 26.09/0.7176 & {\bfseries 29.00/0.8163} \\
\cline{2-4}
& 0.2 & 29.11/0.8157 & {\bfseries 31.94/0.8859} \\
\cline{2-4}
& 0.3 & 31.06/0.8629 & {\bfseries 33.32/0.9086} \\
\hline

\multirow{2}{30pt}{Set14} &  0.1 & 25.23/0.6554 & {\bfseries 27.88/0.7911} \\
\cline{2-4}
& 0.2 & 27.80/0.7534 & {\bfseries 28.58/0.8273} \\
\cline{2-4}
& 0.3 & 29.57/0.8063 & {\bfseries 30.42/0.8867} \\
\hline

\multirow{2}{30pt}{LIVE1} &  0.1 & 25.39/0.6612 & {\bfseries 27.14/0.7478} \\
\cline{2-4}
& 0.2 & 27.58/0.7478 & {\bfseries 28.54/0.8183} \\
\cline{2-4}
& 0.3 & 29.19/0.7991 & {\bfseries 30.07/0.8544} \\
\hline

\multirow{2}{30pt}{General-100} &  0.1 & 26.45/0.7179 & {\bfseries 27.82/0.8048} \\
\cline{2-4}
& 0.2 & 29.88/0.8153 & {\bfseries 30.86/0.8761} \\
\cline{2-4}
& 0.3 & 32.00/0.8639 & {\bfseries 32.88/0.9083} \\
\hline\hline

\end{tabular}
\end{small}
\end{table}

\section{Conclusion}
In this paper, we propose an effective end-to-end compression framework based on two CNNs, one of which is used to produce compact intermediate representation for encoding using an image encoder. The other CNN is used to reconstruct the decoded image with high quality. These two CNNs collaborate each other and are trained using a unified optimization method. Experimental results demonstrate that the proposed compression framework achieves state-of-the-art performance and is much faster than most post-processing algorithms. Our work indicates that the performance of the proposed compression framework can be significantly improved by applying the proposed framework, which will inspire other researchers to design better deep neural networks for image compression along this orientation.


%



\section*{Acknowledgment}
This work is partially funded by the Major State Basic Research Development Program of China (973 Program 2015CB351804), the Science and Technology Commission of China No.17\textrm{-}H863\textrm{-}03\textrm{-}ZT\textrm{-}003\textrm{-}010\textrm{-}01 and the Natural Science Foundation of China under Grant No. 61572155 and 61672188.

\ifCLASSOPTIONcaptionsoff
  \newpage
\fi



\bibliographystyle{IEEEtran}
\bibliography{IEEEabrv,IEEEexample}

\begin{thebibliography}{10}
\providecommand{\url}[1]{#1}
\csname url@samestyle\endcsname
\providecommand{\newblock}{\relax}
\providecommand{\bibinfo}[2]{#2}
\providecommand{\BIBentrySTDinterwordspacing}{\spaceskip=0pt\relax}
\providecommand{\BIBentryALTinterwordstretchfactor}{4}
\providecommand{\BIBentryALTinterwordspacing}{\spaceskip=\fontdimen2\font plus
\BIBentryALTinterwordstretchfactor\fontdimen3\font minus
  \fontdimen4\font\relax}
\providecommand{\BIBforeignlanguage}[2]{{%
\expandafter\ifx\csname l@#1\endcsname\relax
\typeout{** WARNING: IEEEtran.bst: No hyphenation pattern has been}%
\typeout{** loaded for the language `#1'. Using the pattern for}%
\typeout{** the default language instead.}%
\else
\language=\csname l@#1\endcsname
\fi
#2}}
\providecommand{\BIBdecl}{\relax}
\BIBdecl

\bibitem{wallace1992jpeg}
G.~K. Wallace, ``The jpeg still picture compression standard,'' \emph{IEEE
  transactions on consumer electronics}, vol.~38, no.~1, pp. xviii--xxxiv,
  1992.

\bibitem{ghanbari2003standard}
M.~Ghanbari, \emph{Standard codecs: Image compression to advanced video
  coding}.\hskip 1em plus 0.5em minus 0.4em\relax Iet, 2003, no.~49.

\bibitem{zhai2008efficient}
G.~Zhai, W.~Zhang, X.~Yang, W.~Lin, and Y.~Xu, ``Efficient image deblocking
  based on postfiltering in shifted windows,'' \emph{IEEE Transactions on
  Circuits and Systems for Video Technology}, vol.~18, no.~1, pp. 122--126,
  2008.

\bibitem{foi2007pointwise}
A.~Foi, V.~Katkovnik, and K.~Egiazarian, ``Pointwise shape-adaptive dct for
  high-quality denoising and deblocking of grayscale and color images,''
  \emph{IEEE Transactions on Image Processing}, vol.~16, no.~5, pp. 1395--1411,
  2007.

\bibitem{zhang2011image}
R.~Zhang, W.~Ouyang, and W.-K. Cham, ``Image postprocessing by non-local
  kuan’s filter,'' \emph{Journal of Visual Communication and Image
  Representation}, vol.~22, no.~3, pp. 251--262, 2011.

\bibitem{francisco2012generic}
N.~C. Francisco, N.~M. Rodrigues, E.~A. Da~Silva, and S.~M. De~Faria, ``A
  generic post-deblocking filter for block based image compression
  algorithms,'' \emph{Signal Processing: Image Communication}, vol.~27, no.~9,
  pp. 985--997, 2012.

\bibitem{wang2013adaptive}
C.~Wang, J.~Zhou, and S.~Liu, ``Adaptive non-local means filter for image
  deblocking,'' \emph{Signal Processing: Image Communication}, vol.~28, no.~5,
  pp. 522--530, 2013.

\bibitem{zhang2016concolor}
J.~Zhang, R.~Xiong, C.~Zhao, Y.~Zhang, S.~Ma, and W.~Gao, ``Concolor:
  constrained non-convex low-rank model for image deblocking,'' \emph{IEEE
  Transactions on Image Processing}, vol.~25, no.~3, pp. 1246--1259, 2016.

\bibitem{dong2015compression}
C.~Dong, Y.~Deng, C.~Change~Loy, and X.~Tang, ``Compression artifacts reduction
  by a deep convolutional network,'' in \emph{Proceedings of the IEEE
  International Conference on Computer Vision}, 2015, pp. 576--584.

\bibitem{guo2016building}
J.~Guo and H.~Chao, ``Building dual-domain representations for compression
  artifacts reduction,'' in \emph{European Conference on Computer
  Vision}.\hskip 1em plus 0.5em minus 0.4em\relax Springer, 2016, pp. 628--644.

\bibitem{wang2016d3}
Z.~Wang, D.~Liu, S.~Chang, Q.~Ling, Y.~Yang, and T.~S. Huang, ``D3: Deep
  dual-domain based fast restoration of jpeg-compressed images,'' in
  \emph{Proceedings of the IEEE Conference on Computer Vision and Pattern
  Recognition}, 2016, pp. 2764--2772.

\bibitem{yeh2014self}
C.-H. Yeh, L.-W. Kang, Y.-W. Chiou, C.-W. Lin, and S.-J.~F. Jiang,
  ``Self-learning-based post-processing for image/video deblocking via sparse
  representation,'' \emph{Journal of Visual Communication and Image
  Representation}, vol.~25, no.~5, pp. 891--903, 2014.

\bibitem{yoo2014post}
S.~B. Yoo, K.~Choi, and J.~B. Ra, ``Post-processing for blocking artifact
  reduction based on inter-block correlation,'' \emph{IEEE Transactions on
  Multimedia}, vol.~16, no.~6, pp. 1536--1548, 2014.

\bibitem{liu2016data}
X.~Liu, X.~Wu, J.~Zhou, and D.~Zhao, ``Data-driven soft decoding of compressed
  images in dual transform-pixel domain,'' \emph{IEEE Transactions on Image
  Processing}, vol.~25, no.~4, pp. 1649--1659, 2016.

\bibitem{sun2007postprocessing}
D.~Sun and W.-K. Cham, ``Postprocessing of low bit-rate block dct coded images
  based on a fields of experts prior,'' \emph{IEEE Transactions on Image
  Processing}, vol.~16, no.~11, pp. 2743--2751, 2007.

\bibitem{zhang2012reducing}
X.~Zhang, R.~Xiong, S.~Ma, and W.~Gao, ``Reducing blocking artifacts in
  compressed images via transform-domain non-local coefficients estimation,''
  in \emph{2012 IEEE International Conference on Multimedia and Expo}.\hskip
  1em plus 0.5em minus 0.4em\relax IEEE, 2012, pp. 836--841.

\bibitem{zhang2013compression}
X.~Zhang, R.~Xiong, X.~Fan, S.~Ma, and W.~Gao, ``Compression artifact reduction
  by overlapped-block transform coefficient estimation with block similarity,''
  \emph{IEEE transactions on image processing}, vol.~22, no.~12, pp.
  4613--4626, 2013.

\bibitem{he2015deep}
K.~He, X.~Zhang, S.~Ren, and J.~Sun, ``Deep residual learning for image
  recognition,'' \emph{arXiv preprint arXiv:1512.03385}, 2015.

\bibitem{duchi2011adaptive}
J.~Duchi, E.~Hazan, and Y.~Singer, ``Adaptive subgradient methods for online
  learning and stochastic optimization,'' \emph{Journal of Machine Learning
  Research}, vol.~12, no. Jul, pp. 2121--2159, 2011.

\bibitem{zeiler2012adadelta}
M.~D. Zeiler, ``Adadelta: an adaptive learning rate method,'' \emph{arXiv
  preprint arXiv:1212.5701}, 2012.

\bibitem{kingma2014adam}
D.~Kingma and J.~Ba, ``Adam: A method for stochastic optimization,''
  \emph{arXiv preprint arXiv:1412.6980}, 2014.

\bibitem{dong2016image}
C.~Dong, C.~C. Loy, K.~He, and X.~Tang, ``Image super-resolution using deep
  convolutional networks,'' \emph{IEEE transactions on pattern analysis and
  machine intelligence}, vol.~38, no.~2, pp. 295--307, 2016.

\bibitem{kim2015accurate}
J.~Kim, J.~K. Lee, and K.~M. Lee, ``Accurate image super-resolution using very
  deep convolutional networks,'' \emph{arXiv preprint arXiv:1511.04587}, 2015.

\bibitem{kim2015deeply}
------, ``Deeply-recursive convolutional network for image super-resolution,''
  \emph{arXiv preprint arXiv:1511.04491}, 2015.

\bibitem{toderici2015variable}
G.~Toderici, S.~M. O'Malley, S.~J. Hwang, D.~Vincent, D.~Minnen, S.~Baluja,
  M.~Covell, and R.~Sukthankar, ``Variable rate image compression with
  recurrent neural networks,'' \emph{arXiv preprint arXiv:1511.06085}, 2015.

\bibitem{toderici2016full}
G.~Toderici, D.~Vincent, N.~Johnston, S.~J. Hwang, D.~Minnen, J.~Shor, and
  M.~Covell, ``Full resolution image compression with recurrent neural
  networks,'' \emph{arXiv preprint arXiv:1608.05148}, 2016.

\bibitem{theis2017lossy}
L.~Theis, W.~Shi, A.~Cunningham, and F.~Husz{\'a}r, ``Lossy image compression
  with compressive autoencoders,'' \emph{arXiv preprint arXiv:1703.00395},
  2017.

\bibitem{balle2016end}
J.~Ball{\'e}, V.~Laparra, and E.~P. Simoncelli, ``End-to-end optimized image
  compression,'' \emph{arXiv preprint arXiv:1611.01704}, 2016.

\bibitem{li2017learning}
M.~Li, W.~Zuo, S.~Gu, D.~Zhao, and D.~Zhang, ``Learning convolutional networks
  for content-weighted image compression,'' \emph{arXiv preprint
  arXiv:1703.10553}, 2017.

\bibitem{theis2015generative}
L.~Theis and M.~Bethge, ``Generative image modeling using spatial lstms,'' in
  \emph{Advances in Neural Information Processing Systems}, 2015, pp.
  1927--1935.

\bibitem{oord2016pixel}
A.~v.~d. Oord, N.~Kalchbrenner, and K.~Kavukcuoglu, ``Pixel recurrent neural
  networks,'' \emph{arXiv preprint arXiv:1601.06759}, 2016.

\bibitem{krizhevsky2012imagenet}
A.~Krizhevsky, I.~Sutskever, and G.~E. Hinton, ``Imagenet classification with
  deep convolutional neural networks,'' in \emph{Advances in neural information
  processing systems}, 2012, pp. 1097--1105.

\bibitem{ioffe2015batch}
S.~Ioffe and C.~Szegedy, ``Batch normalization: Accelerating deep network
  training by reducing internal covariate shift,'' \emph{arXiv preprint
  arXiv:1502.03167}, 2015.

\bibitem{chang2014reducing}
H.~Chang, M.~K. Ng, and T.~Zeng, ``Reducing artifacts in jpeg decompression via
  a learned dictionary,'' \emph{IEEE transactions on signal processing},
  vol.~62, no.~3, pp. 718--728, 2014.

\bibitem{ren2013image}
J.~Ren, J.~Liu, M.~Li, W.~Bai, and Z.~Guo, ``Image blocking artifacts reduction
  via patch clustering and low-rank minimization,'' in \emph{Data Compression
  Conference (DCC), 2013}.\hskip 1em plus 0.5em minus 0.4em\relax IEEE, 2013,
  pp. 516--516.

\bibitem{dabov2007image}
K.~Dabov, A.~Foi, V.~Katkovnik, and K.~Egiazarian, ``Image denoising by sparse
  3-d transform-domain collaborative filtering,'' \emph{IEEE Transactions on
  image processing}, vol.~16, no.~8, pp. 2080--2095, 2007.

\bibitem{gu2014weighted}
S.~Gu, L.~Zhang, W.~Zuo, and X.~Feng, ``Weighted nuclear norm minimization with
  application to image denoising,'' in \emph{Proceedings of the IEEE Conference
  on Computer Vision and Pattern Recognition}, 2014, pp. 2862--2869.

\bibitem{vedaldi2015matconvnet}
A.~Vedaldi and K.~Lenc, ``Matconvnet: Convolutional neural networks for
  matlab,'' in \emph{Proceedings of the 23rd ACM international conference on
  Multimedia}.\hskip 1em plus 0.5em minus 0.4em\relax ACM, 2015, pp. 689--692.

\bibitem{chen2015trainable}
Y.~Chen and T.~Pock, ``Trainable nonlinear reaction diffusion: A flexible
  framework for fast and effective image restoration,'' \emph{arXiv preprint
  arXiv:1508.02848}, 2015.

\bibitem{he2015delving}
K.~He, X.~Zhang, S.~Ren, and J.~Sun, ``Delving deep into rectifiers: Surpassing
  human-level performance on imagenet classification,'' in \emph{Proceedings of
  the IEEE International Conference on Computer Vision}, 2015, pp. 1026--1034.

\bibitem{sheikh2005live}
H.~R. Sheikh, Z.~Wang, L.~Cormack, and A.~C. Bovik, ``Live image quality
  assessment database release 2,'' 2005.

\bibitem{dong2016accelerating}
C.~Dong, C.~C. Loy, and X.~Tang, ``Accelerating the super-resolution
  convolutional neural network,'' in \emph{European Conference on Computer
  Vision}.\hskip 1em plus 0.5em minus 0.4em\relax Springer, 2016, pp. 391--407.

\end{thebibliography}
%



%
\begin{IEEEbiography}[{\includegraphics[width=1in,height=1.25in,clip,keepaspectratio]{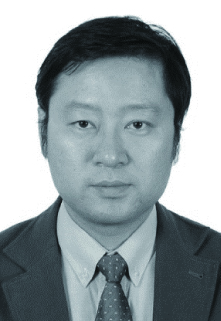}}]{Feng Jiang}
received the B.S., M.S., and Ph.D. degrees in computer science from Harbin Institute of Technology (HIT), Harbin, China, in 2001, 2003, and 2008, respectively. He is now an Associated Professor in the Department of Computer Science, HIT and a visiting scholar in the School of Electrical Engineering, Princeton University.
His research interests include computer vision,  image and video processing and pattern recognition.
\end{IEEEbiography}

\begin{IEEEbiography}[{\includegraphics[width=1in,height=1.25in,clip,keepaspectratio]{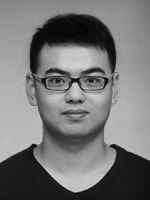}}]{Wen Tao}
received the B.S. degrees in computer science from Harbin Institute of Technology(HIT), Harbin, China, in 2016. He is now working towards the M.S. degree at School of Computer Science and Technology, HIT. His current research interests are in image processing and computer vision.
\end{IEEEbiography}

\begin{IEEEbiography}[{\includegraphics[width=1in,height=1.25in,clip,keepaspectratio]{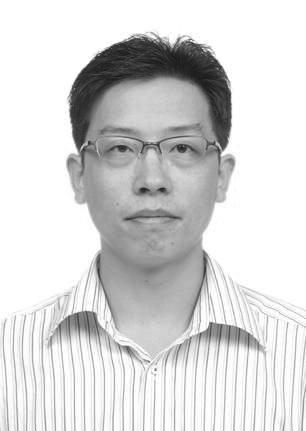}}]{Shaohui Liu}
, Associate professor, has got a Bachelor of Science Degree in Computational Mathematics and Its application Software Development, a Master of Science Degree in Computational Mathematics of Science and a Doctor of Philosophy Degree in Computer Application Technology from Harbin Institute of Technology, Harbin, P. R. China  respectively in 1999, 2001 and 2007. He is a senior member of CCF. Dr. Liu was a visiting professor and post-doctor at Sejong University in South Korea, and a Visiting Scholar at University of Missouri Columbia in America. His research mainly includes image and video processing and analysis, computer vision, multimedia security. In the related fields, he has coauthored more than 80 papers which are cited more than 1000 times totally.
\end{IEEEbiography}

\begin{IEEEbiography}[{\includegraphics[width=1in,height=1.25in,clip,keepaspectratio]{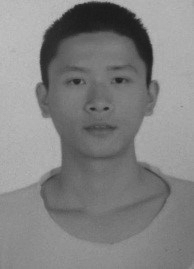}}]{Jie Ren}
received his B.S. degree in 2015 from University of Electronic Science and Technology of China; now, he is a master in Harbin Institute of Technology. His main research interest includes image processing and multimedia compression technology.
\end{IEEEbiography}

\begin{IEEEbiography}[{\includegraphics[width=1in,height=1.25in,clip,keepaspectratio]{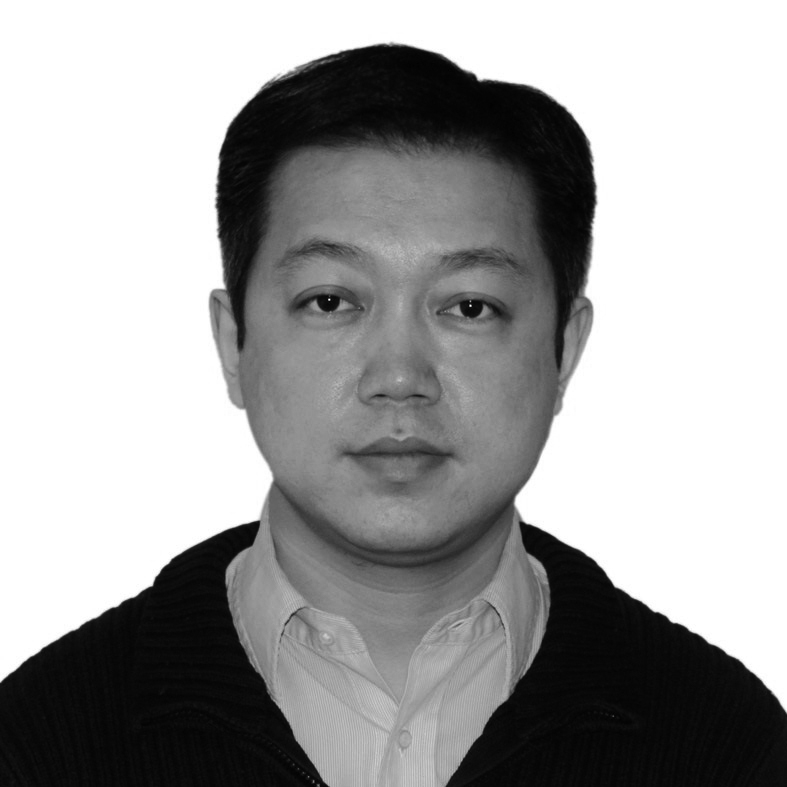}}]{Xun Guo}
received his Ph.D. degree in computer science from Harbin Institute of Technology, China, in 2007. From 2007 to 2012, he was with MediaTek Inc. as group manager, where he led a research team and worked mostly on video compression, especially on technology development for HEVC standard. He joined Microsoft Research Asia in 2012, where he is now a lead researcher. His research interests include video coding and processing, multimedia system and computer vision.
\end{IEEEbiography}

\begin{IEEEbiography}[{\includegraphics[width=1in,height=1.25in,clip,keepaspectratio]{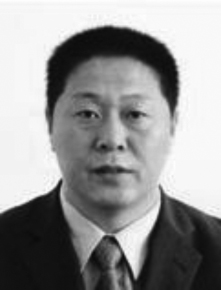}}]{Debin Zhao}
(M'11) received the B.S., M.S., and Ph.D. degrees in computer science from Harbin Institute of Technology (HIT), Harbin, China, in 1985, 1988, and 1998, respectively.
He is now a Professor in the Department of Computer Science, HIT. He has published over 200 technical articles in refereed journals and conference proceedings in the areas of image and video coding, video processing, video streaming and transmission, and pattern recognition.
\end{IEEEbiography}







\end{document}